\documentclass{llncs}
\usepackage[nocompress,nospace]{cite}
\usepackage{amsmath}
\usepackage{array}
\usepackage{url}
\usepackage{color}
\usepackage{listings}
\usepackage{caption}
\usepackage{subcaption}
\usepackage[pdftex]{graphicx}
\graphicspath{{./img/}, {./img/figures/}}
\DeclareGraphicsExtensions{.pdf,.jpeg,.png,.svg}

\begin{document}
\title{Traceability of Deep Neural Networks}

\author{Vincent Aravantinos \and Frederik Diehl}
\institute{fortiss GmbH\\
  Guerickestr. 25, 80805 Munich, Germany\\
  \email{\{aravantinos,diehl\}@fortiss.org}}

\maketitle

\begin{abstract}
  \emph{[Context.]} The success of deep learning makes its usage tempting in safety-critical applications.
  However the development of such applications typically has requirements
  which do not allow the usage of machine learning,
  in particular \emph{requirements traceability} of software artifacts.
  \emph{[Problem.]}
  Code constituting a neural network is not comparable to classical code
  and requirements for applications where neural networks are needed typically cannot be
  concretized into low-level requirements.
  \emph{[Proposed solution.]}
  We investigate the neural network equivalents of code and low-level requirements
  and propose various traces that one could consider as a replacement for classical notions.
  We also propose a form of traceability in order to deal with the particular
  trial-and-error development process for deep learning.
\end{abstract}

\keywords{Traceability \and Neural networks \and Deep learning \and Safety}

\section{Introduction}

The success of deep learning (DL), in particular in computer vision,
makes its usage more and more tempting in many applications, including safety-critical ones.
However the development of such applications must follow standards (e.g., DO178 \cite{do178}, ISO26262 \cite{iso26262}) which typically 
do not envision the usage of machine learning.
At the moment, practitioners therefore cannot use machine learning for safety-critical functions
(e.g., ASIL-D for ISO26262, or DAL-A for DO178).

There exist various attempts to address this issue whether in
standardization committees (e.g., ISO/IEC JTC 1/SC 42 or DKE/DIN \cite{DKE})
or in the academic community 
(various initiatives towards explainable AI, e.g., \cite{chih-hong-metrics}),
but they are all far from mature and definitely not usable as of today
or do not really address the problem:
most standardization approaches just try to map one-to-one classical software 
engineering processes like the V-model to deep learning.
Furthermore, no current academic approach provides a completely satisfying solution to the lack of 
understandability of deep neural networks (DNN).

In this paper, we try to find a pragmatic approach, which focuses on artifacts rather than on processes:
we are not prescriptive regarding the activities which produce these artifacts.
More precisely, we focus only on artifacts which are worth being identified
during the development of DNNs \emph{for the sake of traceability.}
Consequently, this paper does not provide a ready-made solution, which a practitioner could follow one-to-one.
However, it provides concrete descriptions which could be sufficient to provide a first guidance.

We restrict the scope of this paper to the following:%
\begin{itemize}
  \item DNNs for supervised learning (no reinforcement learning,
    no unsupervised learning).
  \item We focus only on software, not on system: traces from software requirements to system requirements 
    are out of scope, as is item-level hazard analysis.
  \item We do not focus on binary code or deployment thereof on hardware platform.
  \item We assume a fixed, non-evolving, dataset. This does not comply with most real cases in, say, 
    autonomous driving, where data is continuously collected.
    Even if not continuously collected, the dataset has so much influence on the training that one can hardly ignore its evolutions for proper traceability.
    Still, there are already sufficiently many questions to address without considering this evolution, which is why we leave this out of focus in this paper.
  \item We focus on \emph{functional} requirements.
\end{itemize}
Lifting these restrictions is left to future work.

The rest of the paper is organized as follows:
Section \ref{sec:related} presents related work.
Section \ref{sec:traceability} recalls the concept of traceability.
Section \ref{sec:DL} provides a traceability-amenable presentation of deep learning.
Section \ref{sec:tracing-dl} contains the main contribution of this paper:
it analyzes which DNN artifacts replace classical software artifacts
and suggests new artifacts to enable the traceability of DNNs.
Section \ref{sec:future-work} concludes the paper and lists open questions for future research.

\section{Related work}
\label{sec:related}

The usage of DNNs to support safety-critical functions is commonly recognized
as a huge challenge \cite{chih-hong-challenges}.
There are attempts towards the certification, verification, or explainability of DNNs,
of which we provide now a short overview.
None of them however addresses the traceability of DNNs.

Back in 1996 a set of requirements for a standard certifying the usage of neural networks
in safety critical applications was gathered \cite{requirements4standard}.
Traceability is (indirectly) mentioned as a problem to address, but no solution is provided.

In 1999, the classical waterfall model was adapted to the development of DNNs \cite{process4NN}.
Even though the problematic of traceability (in particular to the data)
is indirectly addressed, the proposed process is very much oriented towards \emph{activities}
described in an informal manner, rather than on a concrete list of artifacts to provide.
The resulting process remains thus very high-level:
there is a large freedom of interpretation about what the artifacts shall actually contain,
which therefore does not provide enough information to trace DNNs.

There has been attempts to apply principles of software engineering
to NNs \cite{engineering4NN},
in particular to address the lack of reproducibility in the development of NNs.
Even though the terminology and techniques have changed a lot since 2004,
the identified problems are still relevant.
Yet, the solutions of the paper do not answer the need for traceability
and seem to hardly match nowadays' practice.

The discrepancy between the recommendations of the ISO 26262 and the methods 
actually used in practice has been analyzed \cite{ISO26262-vs-NN}.
This cannot be directly used for traceability but is indirectly a very useful source of information.

A safety argumentation for the usage of DNNs,
formalizing it partly using GSN has been developed \cite{Kurd-kelly}.
The abstraction level of this work is higher than what we address in this work,
meaning in particular that we cannot derive directly from it any notion of trace.
The same applies for more recent work also applying GSN in the context of NNs
\cite{burton-case,gauerhof-case}.

There are lots of heuristics, lessons learnt, best practices, all of which are available 
from informal sources on the internet\footnote{\url{https://developers.google.com/machine-learning/rules-of-ml/}
, \url{https://blog.slavv.com/37-reasons-why-your-neural-network-is-not-working-4020854bd607}}
or analyses regarding the technical debt
of machine learning \cite{technical-debt-NN}.
None provide a concrete notion of traceability but they
help to analyze which information is worth tracing.

Interpretability and explainability of DNNs
are very hot topics, with few actual solutions at the moment,
the most advanced ones provide rather hints of solutions \cite{building-blocks-interpretability}, 
Various such solutions have been gathered into a set 
of metrics to use in order to assess a network \cite{chih-hong-metrics}.
None of these are entirely satisfying nor do they relate directly to traceability.
However they all could provide technical insights in order to assess what should be traced or not.

Finally, there has been attempts to set grounds for a ``rigorous science'' of machine learning \cite{rigorous-science}:
again very useful, but not related to traceability.
Many safety-related problems have been identified for AI, especially for reinforcement learning \cite{safeAI}.
The identified challenges and corresponding attempts of solutions are 
a potential source of inspiration to identify sources of problems and
thus to analyze whether those sources can be tackled \emph{via} traceability.

\section{Preliminary: Traceability}
\label{sec:traceability}

\subsubsection{Traces and artifacts.}
\begin{figure}[!ht]
  \centering
  \begin{subfigure}[b]{4.5cm}
    \footnotesize
    \begin{lstlisting}[xleftmargin=-5ex]
    1 //f_456 takes:
    2 // - x: ...
    4 // It returns ...
    5 //
    6 // [REQ_123]
    7 int f_456(int x) {
    8 ...
    9 }
    \end{lstlisting}
    \caption{Trace}
    \label{fig:trace}
  \end{subfigure}
  \begin{subfigure}[b]{7.5cm}
    \includegraphics[width=7.5cm]{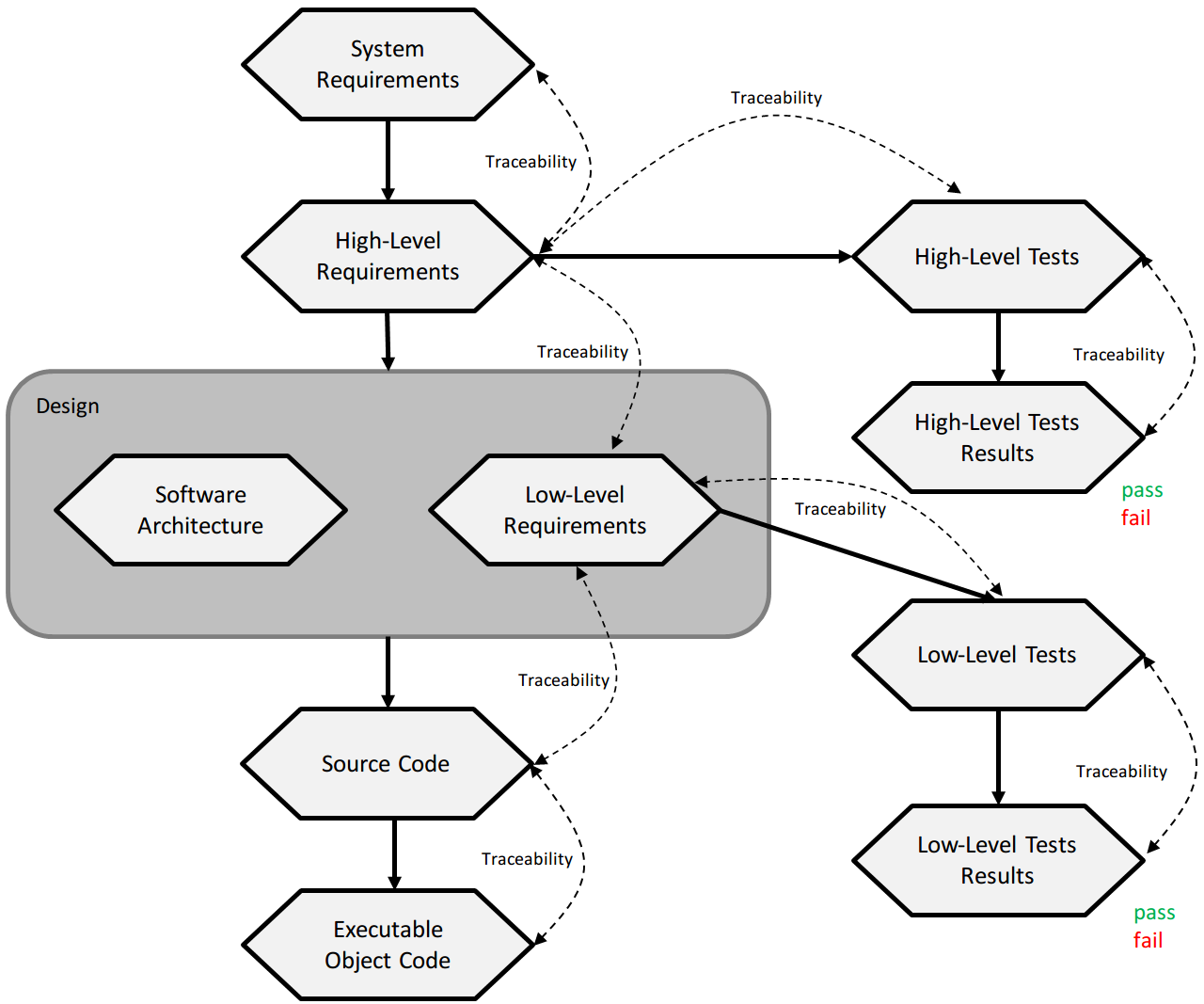}
    \caption{Classical artifacts -- inspired by the DO333 \cite{do333}}
    \label{fig:classical-structure}
  \end{subfigure}
  \caption{Traces and artifacts}
\end{figure}

We call any document delivered together with the software an \emph{artifact}.
This includes executable code, source code, test results, development plans
and are usually defined by a standard,
like ISO26262 \cite{iso26262} or DO178C \cite{do178}).
The delivered artifacts generally have dependencies between each other:
typically, the source code should fulfill requirements,
software requirements should refine system requirements,
executable code derives from source code.
Keeping these dependencies implicit increases the risk that a dependency be wrong or forgotten.
\emph{Traces are a concrete attempt to make these dependencies explicit.}
Every pair of artifacts is in principle subject to being traced from/to each other,
but in this paper we consider especially traces from source code to requirements.
Consider for instance a requirement of ID, say, $REQ\_123$
and a piece of code defining a function, say, $f\_456$ (see Fig.~\ref{fig:trace}).
Then a typical trace is nothing more than a comment just before the function
simply stating $[REQ\_123]$ (see line 6 in Fig.~\ref{fig:trace}).
Writing down a trace is generally a \emph{manual} activity:
engineers look up the code and requirements and add manually the comment above.

\subsubsection{High- and low-level requirements.}
In many cases, requirements are not concrete or precise enough to be traced directly from the source code.
One therefore first refines the requirements into low-level requirements (LLR), 
which can be traced from the code.
The former requirements are then called high-level requirements (HLR).
As an example, the DO178C standard names define LLRs as follows:
``Low-level requirements are software requirements
from which Source Code can be directly implemented without further information.'' \cite{do178}.
LLR should themselves be traced to HLR in order to have complete traceability.%

Refining HLR into LLR goes hand in hand with architectural decisions:
the requirements can be decomposed only once the function is decomposed into smaller 
functions, to which one assigns more concrete requirements.
This is why the DO178C refines the HLR into two artifacts:
the LLRs on one hand, and the Software Architecture on the other hand.
The software architecture defines a set of components
and connections between these components 
(or, more precisely at that stage, the \emph{interfaces} of such components).
In the following, we will call the definition of such interfaces \emph{interface requirements}
(even if, in practice, they are not necessarily explicitly defined as requirements).
The LLRs can then be mapped to each interface.
Finally, the LLR and the software architecture should be the only
information necessary to write down the source code.

Fig. \ref{fig:classical-structure} represents the artifacts mentioned above 
(and others, especially related to verification).
Of course, every artifact shall be traced to the artifact(s) it refines, typically in a bi-directional manner: 
every piece of information found in a refined artifact shall be found in the corresponding refining artifact,
and conversely, every piece of information -- except design decisions -- found in a refining artifact shall be found in the refined one.
In the DO178, the software architecture is not traced back to the HLR because it is a design decision.

\subsubsection{Rationale.}

Understanding the rationale behind traces
(1) enables to understand why it is challenging to trace DNNs,
and (2) gives hints to investigate relevant alternatives to classical traces.
A trace serves the purpose of ensuring that a piece of code is justified by a requirement,
without being a structured or formal justification.
Actually, traces can be seen as a pragmatic approach to support what
one would actually like to express as a proper \emph{argumentation}.
They at least enforce that people \emph{think} about a justification.
In fact, traceability does enable to identify sources of error:
when engineers fail to trace a piece of code
then they might discover that it is not necessary or, even worse,
that it introduces unwanted functionality.
Conversely, if a requirement is not traced back by to any code,
then it is often a hint that its implementation was forgotten.
For the same reason, traceability is a tool for assessors 
to detect potential pitfalls during development.
This is what is illustrated in Fig.~\ref{fig:classical-structure} by the bi-directional arrows for traceability:
having traces syntactically on each side is easy; 
it is however harder to ensure \emph{coverage} of traceability on both sides,
e.g., \emph{all} HLR are traced to some LLR and \emph{all} LLR are traced back
to some HLR (the latter typically does not 
happen since some LLRs depend on design decisions).

Some standards like, e.g., DO178C, do not impose an order on how artifacts shall be developed.
For instance, even though code shall be traced to requirements,
it does not mean that one is forced to follow a waterfall model:
one might just as well work iteratively, define requirements,
then code, then go back to requirements, etc.
An important point of traceability is that, no matter how one reached the final state
of development (e.g., iteratively or waterfall), it should be possible to
justify that this final state is coherent.
Consequently, one might very well develop all the artifacts without traceability,
and only at the end develop the traces (of course this is not recommended).
This is why we emphasized in introduction that this paper is not process- but artifact-oriented:
we do not impose \emph{how} engineers should work but only \emph{what} they should deliver.

\section{Deep learning artifacts}
\label{sec:DL}

This section presents concepts and terms related to deep learning, in a way that makes them 
amenable to comparison with the artifacts of the previous section.

\begin{figure}[!ht]
  \centering
  \begin{subfigure}[t]{4cm}
    \includegraphics[width=4cm]{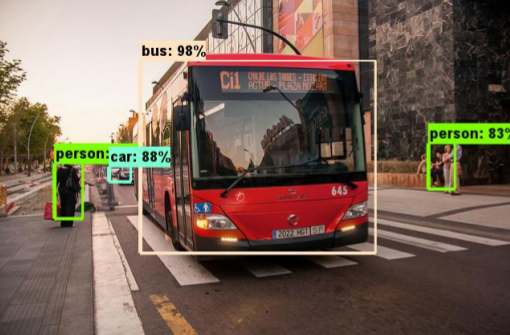}
    \caption{An example from an object recognition dataset, with bounding boxes and types
      of objects plotted into the image (from \cite{mobilenet})}
    \label{fig:bounding-boxes}
  \end{subfigure}
  \begin{subfigure}[t]{8cm}
    \includegraphics[width=8cm]{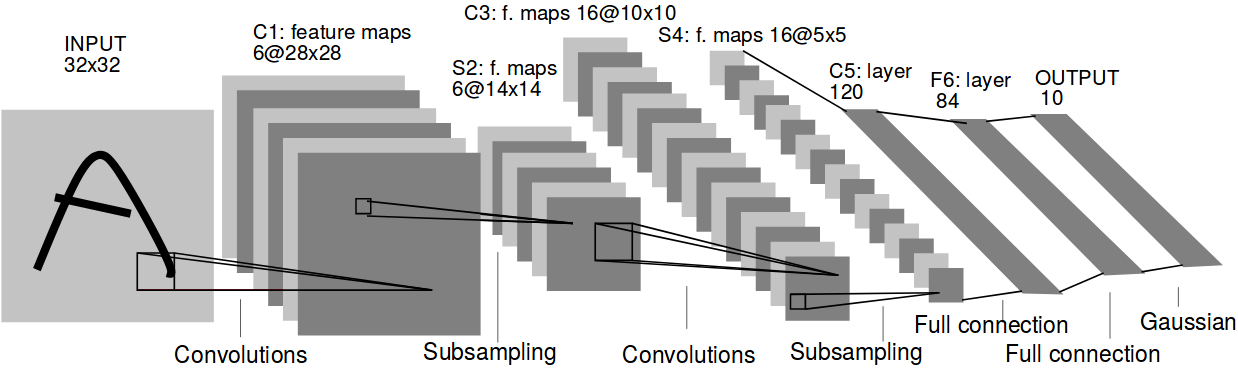}
    \caption{Architecture of LeNet, an early and well-known DNN architecture to classify hand-written characters
      (from \cite{lenet})}
    \label{fig:arch}
  \end{subfigure}
  \caption{Bounding boxes and DNN architecture}
\end{figure}

To implement a required function using a DNN, one needs:
\begin{itemize}
  \item A \emph{dataset} containing both instances of the input, e.g., images,
    and of the output, e.g., annotations denoting bounding boxes for pedestrians' positions
    (see, e.g., Fig. \ref{fig:bounding-boxes}).
    In the following, we will consider both separately:
    the \emph{raw dataset} for the input,
    and the \emph{labels} for the corresponding expected output.

  \item A neural network \emph{architecture}.
    A DNN is a function typically structured as a sequence of parallel operations transforming the input.
    This sequence of operations has the form of a layer-based architecture.
    It plays an essential role in distinguishing competing networks
    and results from the ingenuity of the deep learning engineer.
    See Fig. \ref{fig:arch} for a classical example.

  \item A \emph{loss function}.
    During the learning process, if a DNN predicts a value that turns out to be wrong,
    it does not simply change its weights so that
    it gets suddenly right, rather it tries to get \emph{closer} to the solution.
    There are various reasons for that, but all that matters for this paper
    is that engineers shall define ``closer'':
    to do so, they define a socalled \emph{loss function} providing a positive real
    number telling \emph{how wrong} the system is.
    The corresponding code influences the learning but does not go into the final product.
\end{itemize}
All of these are provided to a DL framework which takes care of generating a trained network
(e.g., TensorFlow\footnote{https://www.tensorflow.org/} or PyTorch\footnote{https://pytorch.org/}).
It is essential not just to understand
the artifacts themselves, but how they are developed.
Typically, the sequence of decisions are as follows:
\begin{enumerate}
  \item Collect data and preprocess it: 
    re-shape the information, fix missing values, extract features.
    \textbf{Delivered artifacts:} raw dataset, preprocessing functions.

  \item Annotate the raw data.
    \textbf{Delivered artifact:} labelled dataset.

  \item Split the labelled dataset into training, validation and testing datasets.
    The difference between the two latter is that one uses the evaluation on the validation dataset
    in order to improve the DNN design,
    whereas the testing dataset is used only once no more iteration is planned,
    to assess the final quality of the DNN.%
    \footnote{These terms are here used a bit opposite to their classical usage in a safety context.}
    \textbf{Delivered artifacts:} train, validation and test datasets.

  \item Design the architecture.
    \textbf{Delivered artifact:} DNN architecture.

  \item Define the ``learning configuration'' (non-standard term),
    i.e., all remaining aspects influencing the learning process:
    loss, learning parameters (e.g., dropout, learning rate, maximum steps), hyper-parameters,
    or, at lower level: random seeds, versions of software and hardware used for training.
    Overall, the configuration shall be the minimal piece of information so that,
    together with the training set and the architecture, they characterize uniquely the learned DNN.
    \textbf{Delivered artifact:} 
    Typically scattered across different artifacts:
    e.g., parameters stored in code, loss having its own source file, etc.
    Ideally, this should be gathered in centralized files \cite{sacred}.

  \item Train the DNN architecture.
    \textbf{Delivered artifact:} (trained) weight values.%
    \footnote{This artifact is memory rather code:
      training does not alter the code itself, but only the memory values of the variables \emph{used} by the code.}

  \item Post-process the trained DNN (if necessary):
    many learning strategies require a change between learning and inference phase 
    (e.g., dropout is applied only during learning).
    \textbf{Delivered artifact:} inference architecture.%
    \footnote{Here, on the opposite, the code theoretically changes but not the weights.
      In practice, though, the code remains a same and the switch between training and inference mode 
      is controlled by a simple boolean flag, so there is no need for separate artifacts.}

  \item Test the resulting DNN on the validation dataset.
    \textbf{Delivered artifact:} test results (e.g., real number representing accuracy).

  \item Change the architecture or the learn configuration (4--5) based on the results
    and repeat steps 6--9 until the targeted objectives are reached.

  \item Assess the quality with the test set.
    \textbf{Delivered artifact:} final metric value.

  \item Depending on the used framework, serialize/export the network in order to use it in production, 
    e.g., to be linked from a C++ source file, and compile it.
    \textbf{Delivered artifact:} executable code usable in production.
\end{enumerate}
Similarly to code, the produced DNN is obtained by trial-and-error.
Contrarily to code however, the resulting DNN typically cannot be understood 
without looking at the changes which led to it.
Often, it is even the only way to understand why a DNN is finally obtained.
This has of course a big impact on justifiability of a DNN and therefore on traceability,
as we will see in Section \ref{sec:trial-and-error}.

\section{Tracing DL artifacts}
\label{sec:tracing-dl}

To devise traces for DNNs, it is helpful to map classical development (Section \ref{sec:traceability})
with DL development (Section \ref{sec:DL}).
To do so, we analyze, for all artifacts of Fig.~\ref{fig:classical-structure},
which DL artifact plays a similar role.
Some artifacts are easy:
\textbf{System requirements} and \textbf{HLR} are present independently of using DL or not 
--- either way, one needs requirements both at the system level and for the software.
The same applies to \textbf{Executable object code}: either way, the final product is code.

Much more difficult is the case of \textbf{Software architecture} and \textbf{LLR}:
\emph{Our position is that HLRs that are implemented with DNNs generally
cannot be refined into a software architecture and an LLR.}
Indeed, if this were the case, then engineers could themselves decompose the problem 
into subproblems, which could in turn be manually implemented.
However, DNNs are precisely used when no such engineered decomposition is successful;
this is in particular the case in computer vision, where decades of human engineering were
outperformed by DNNs, which precisely came up ``themselves'' with more relevant decompositions.
Note that this is not particularly a property of DNNs per se, but rather of the \emph{functions} 
for which it makes sense to use DNNs.
As mentioned earlier, according to DO178C, the LLR shall contain sufficient information to write down the code.
We therefore suggest to substitute LLR and software architecture with all the information,
which suffices for the \emph{DL framework} to generate code:
\textbf{Training dataset}, \textbf{DNN architecture} and \textbf{learning configuration}.

Regarding \textbf{Source code}, a simple position is to consider that there is still source code at the end.
However, as mentioned in Section \ref{sec:DL}, the outcome of the DL framework is rather memory values than code:
the code itself is still code, which was written by the DL engineer to specify the architecture.
We suggest therefore to split the source code block into the part that is manually written and the part 
that is generated: \textbf{Inference architecture} 
(which, as mentioned earlier, need not be different than the DNN architecture)
and \textbf{Weight values}.

\begin{figure}[!ht]
  \begin{subfigure}[b]{6cm}
    \centering
    \includegraphics[width=6cm]{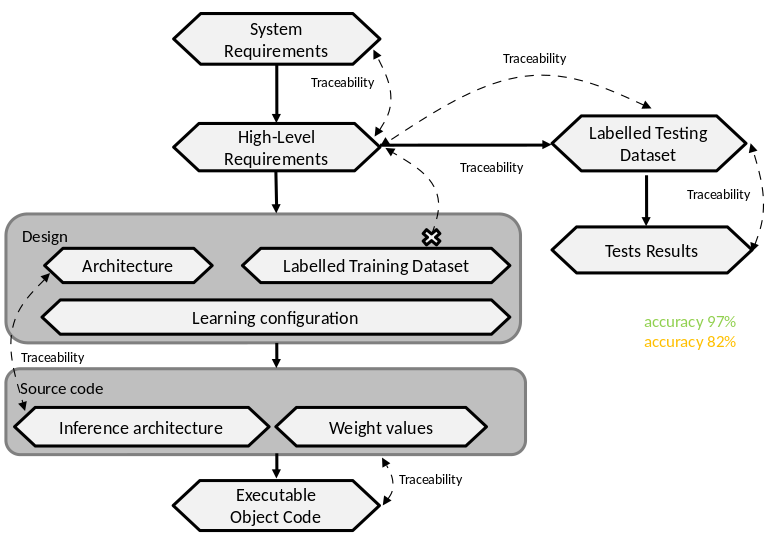}
    \caption{DNN artifacts}
    \label{fig:DNN-structure}
  \end{subfigure}
  \begin{subfigure}[b]{6cm}
    \centering
    \includegraphics[width=6cm]{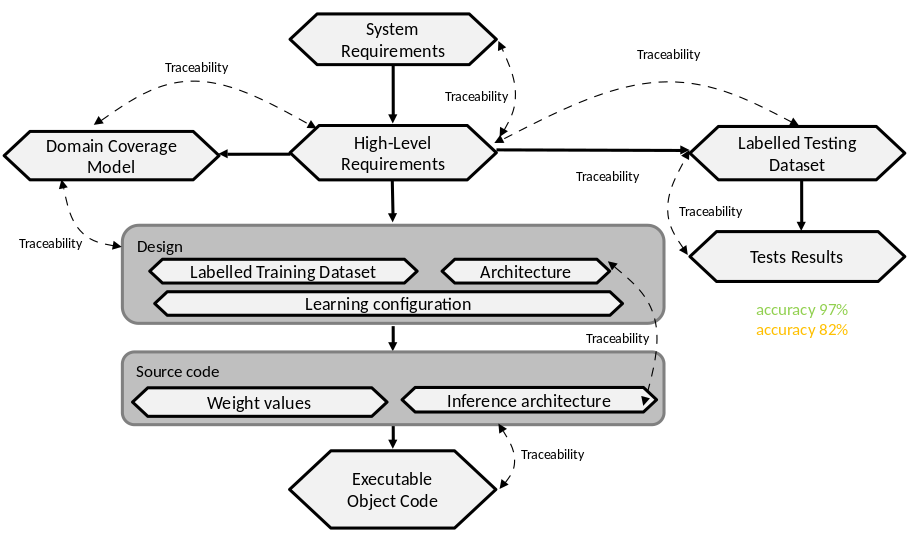}
    \caption{Integration of domain coverage model}
    \label{fig:DNN-structure-domain}
  \end{subfigure}
  \caption{Adaptations of Fig. \ref{fig:classical-structure} for DNN}
\end{figure}

The result yields Fig. \ref{fig:DNN-structure}.
Note that high-level tests are now replaced by the testing/validation set: the name differs but the role is the same.
An important difference is that, contrarily to classical test, one cannot expect 100\% passed tests:
it is in the nature of the functions solved using NNs that they are not perfect.
Low-level tests disappear since there are no more LLR.

We consider traces between these artifacts,
more precisely between:
HLR and training dataset (resp. learning configuration, resp. architecture),
source code and training dataset (resp. learning configuration, resp. architecture).
Some are simple: e.g.,
inference architecture to learning architecture (when not the same anyway) and to no other design artifact,
or between the training dataset version and the learnt weights
(indeed, it is easy in practice to lose track of which version of a dataset was used to train a given network).%
\footnote{Beyond the scope of this paper but also essential is the traceability
  between a dataset and its \emph{sources}, e.g., sensors calibration setup or sensor driver versions.}

Tracing other artifacts is much more challenging and the next sections are dedicated to this problem:
the next section deals with traces between HLR and training dataset,
the following section deals with all other traces.

\subsection{Traceability between HLR and training dataset}
\label{sec:traceability-dataset-hlr}

Traces between HLR and dataset may seem simple:
one just needs to trace every element of the dataset to the HLR.
Consider, e.g., the HLR ``The function shall recognize obstacles in urban context''.
In principle, it is simple to trace the dataset to such requirements:
e.g., pictures in the dataset taken in urban context shall be traced to the corresponding requirement.
However, the sort of information usually found in an HLR often applies uniformly 
to \emph{all} elements of the dataset:
e.g., if the function shall work only in urban context then \emph{all} images of the dataset will be urban.
This would entail tracing the \emph{entire} dataset to the HLR, which would be so general
that it would not really support the rationale of tracing:
it would not provide any form of justification of this particular dataset.
Instead, one expects every datum to be justified and thus traced \emph{individually}.

\subsubsection{Domain coverage model.}
In practice, HLR are not refined enough to allow such a fine-granular traceability.
Instead, we propose a new artifact between HLR and dataset: the ``domain coverage model''.
This model shall provide a decomposition of the domain of the HLR in such a way that various elements of the dataset
can be traced to different \emph{components} of this decomposition.
For instance, for the example above, ``urban'' is not enough:
one should detail the different forms of environment that are encountered:
e.g., ``one-way street'', ``roundabout'', etc.
(in that case, the model connects strongly with the Operational Design Domain -- ODD --
but it needs not be the case if the function to be performed by the DNN 
does not directly work with data coming from the sensors).
Every piece of data matching the corresponding part of the domain model shall be then traced to it.
These parts should be themselves traced towards the HLR.%
\footnote{This might be a useful tool to identify misunderstandings regarding the environment,
  e.g., is a portion of \emph{highway} which is \emph{within the limits of a city} considered urban?}

If working in a very structured context, e.g., where model-based requirements engineering is used \cite{MbRE},
the domain coverage model could be formalized to some extent.
In such cases, coupling it with the prior definition of a coverage criterion
comes in close connection with coverage-driven verification \cite{cdv}
or model-based testing \cite{MbT}.
The difference with these classical approaches is merely the size and diversity of the model, 
which is much bigger in DL than in classical engineering.
Similar approaches have been proposed for machine learning in the literature, 
see e.g., \cite{dataset-testing}, but to a smaller and less systematic extent.
The domain model is a similar to a plant in control engineering:
it models the environment, but without its dynamics.
Fig. \ref{fig:DNN-structure-domain} reflects the new artifact and the corresponding traceability.

Note that, even though the discussion above targets especially the raw dataset,
the same applies to the labels if their domain is complex enough:
for instance, if the DNN shall provide the position on a pedestrian, then it is important
to ensure that the domain of positions is adequately covered.

\subsubsection{How detailed is detailed enough?}
It is typically very hard for a requirement engineer to know \emph{beforehand}
which level of granularity to put in such an domain coverage model.
Actually the level of granularity probably depends on the dataset itself,
and can thus be identified only once the dataset is already (or at least partially) present:
this is counter-intuitive regarding the usual notion of requirement.
However, as mentioned, we do not focus on the order in which artifacts are delivered
but only on ensuring their mutual consistency.
In this respect, it is acceptable to generate or modify \emph{a posteriori} such a requirement.

To find out the proper level of granularity, one shall keep in mind that
such a domain coverage model shall serve as a \emph{tool}
to analyze the dataset by justifying why a particular datum is in it
and identifying cases where some situation might not be covered.
Consequently, if \emph{too many} pieces of data trace to the same domain model part,
then this part is probably defined with a too high-level of granularity and traceability is useless.
Conversely, if \emph{too few} pieces of data trace to one domain model part,
then this part is probably defined with a too low-level of granularity 
and coverage will not be reachable as displayed in Fig. \ref{fig:DNN-structure}:
the traceability arrow between HLR and dataset is not bidirectional.

Defining ``too many'' or ``too few'' is beyond the scope of this paper,
but should of course be defined in a rigorous manner depending on the context.
If this approach turns out convincing, we actually suspect that coverage domain modelling
will become a non-trivial engineering phase of its own.

\subsection{Trial-and-error traces}
\label{sec:trial-and-error}

All remaining traces relate to the main DL artifacts:
weights, learning configuration, architecture, training dataset.
A na\"ive solution consists in tracing \emph{weights} to dataset elements:
this can be performed automatically by the DL framework by tracing to the dataset elements,
which influence a given weight ``the most'' \cite{chih-hong-challenges}.
Even though doable in theory, this approach is irrelevant in practice,
it is acknowledged as impossible -- at least as of today --
to interpret the role of one particular neuron.
This sort of traces will thus not fulfill the traceability rationale:
if a reviewer inspects the involved artifacts in their state at the end of the project,
they will not understand them nor their connection to previous artifacts.
Note that this is a problem classically encountered with code generators,
which cannot be trusted without any argumentation.
Unlike ``classically'' generated code though, DNNs cannot be understood by manual inspection.

\subsubsection{Trial and error.}
\begin{figure}[!ht]
  \centering
  \begin{subfigure}[b]{5cm}
    \centering
    \includegraphics[width=5cm]{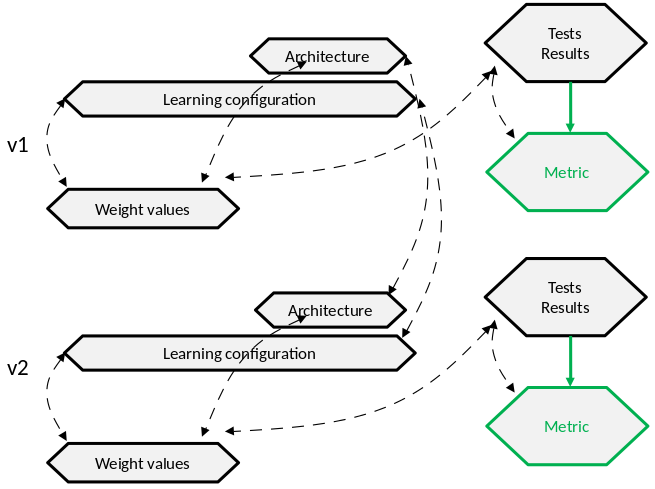}
    \caption{Trial-and-error}
    \label{fig:trial-and-error}
  \end{subfigure}
  \begin{subfigure}[b]{7cm}
    \centering
    \includegraphics[width=7cm]{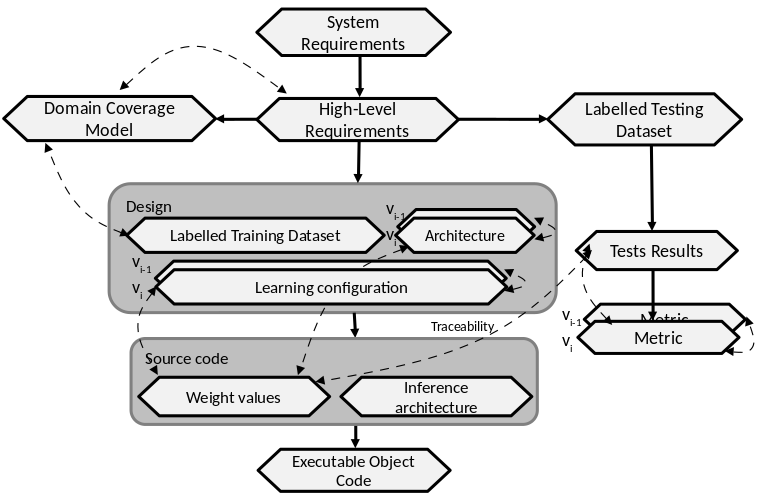}
    \caption{Corresponding extension of Fig. \ref{fig:DNN-structure-domain}}
    \label{fig:trial-and-error-process}
  \end{subfigure}
  \caption{Trial and error}
\end{figure}

Instead of waiting for explainable AI to provide solutions,
we suggest in this paper to trace the engineers' decisions instead of the artifacts themselves:
if artifacts are not understandable, engineers' decisions shall be.
How do engineers come up with architectures or learning configurations?
They essentially \emph{try} them, test them, and try again until they cannot improve the results anymore.
In other words, these decisions are intrinsically based on trial-and-error:
see Fig. \ref{fig:trial-and-error} for an illustration.
In classical software engineering, trial and error is not considered for traceability,
it is actually rather the opposite:
one expects from traceability that we can ensure the coherence of the artifacts
\emph{in their final state}, i.e., independently of how they were obtained,
by trial-and-error or not.
However, DNN development relies so intrinsically on trial-and-error,
that we feel necessary to embrace this kind of activity even for traceability.
Future developments might provide more predictable and reproducible approaches 
to the development of DNNs, in which case the approach of the present section will become obsolete.
At the moment this paper is written, the present approach
shall be considered as a pragmatic attempt that can be applied today.

In case of trial-and-error, the only justification that one can provide
is that a given artifact is \emph{better} than its previous version.
We thus propose to trace every new artifact obtained
by trial-and-error to its previous version.
This requires storing not only the final artifact but
also all the previous versions of it, which is done already in a standard manner
in the context of configuration management and using version control.
Pairing configuration management with traceability
forces the engineer to do more than just tagging 
a new version in their version control system:
they must also think about the \emph{justification} of the new increment.

\subsubsection{Metric.}
To apply this approach, it is essential to define what ``better'' means,
we introduce therefore an additional \emph{metric} artifact
to measure the quality of the inference DNN -- which engineers anyways normally do.
Such a metric should be defined according to the function
(e.g., we allow a car to be mistaken for a pedestrian, but not the other way around)
and can range from simple cases like accuracy and/or recall to complex
combinations of functions \cite{chih-hong-metrics}.
There should be only one metric, but there are several metric \emph{values}.
These values shall be traced to the weight values and inference architecture with which these were obtained.
\emph{The essential addition is then to require that every version of the network which is obtained by increment
of a previous one shall be traced to the metric value obtained with this previous version:
one can then easily check if the new value indeed is an improvement.}
If the metric changes during the project or is defined only a posteriori,
then one needs to re-check the \emph{entire trial-and-error chain} leading to the final version of the DNN.
Fig. \ref{fig:trial-and-error-process} summarizes these additions.

This process is a big hindrance for the practitioner, but:
(1) not providing an argumentation for a claimed improvement is recognized
as a problem by the DL community itself (see ``Explanation vs Speculation'' in
\cite{troubling}),
and (2) it is much easier to apply than any approach currently taken in the field of explainable AI.

\subsubsection{Limitations.}
In its current state, our proposition can be ``tricked'':
a developer can deliver only their last trial and claim that it was, by chance, the first one.
A way to circumvent this, is to impose some restrictions on the first delivered version,
e.g., imposing that it belong to a catalogue of authorized ``primitive'' DNNs.
A developer cannot then just deliver immediately a complex DNN without tracing it to a previous primitive one.
This still would not be enough:
once again, a developer can silent all the intermediate developed versions 
and trace directly from the last to the first version,
which goes against the intent of our approach.
To circumvent this, one could restrict allowed increments:
e.g., adding only one layer at a time, having a default size for layers, etc.
This might be too restrictive in practice:
some DNNs only show their benefits after having added a certain number of layers,
but all the smaller versions with less layers are all equally bad.
Investigating such restrictions goes beyond the present paper.

\section{Conclusion and open questions}
\label{sec:future-work}

We addressed the traceability of DNNs in a pragmatic manner:
we analyzed the parallels and differences between DNNs and classical software development,
and proposed accordingly adaptations of the notion of trace for DNNs.
Instead of blindly mapping classical software activities to DL activities,
which would lead to mismatches with the practice of DL,
we tried to embrace some specificities of ``real-life'' DL, in particular trial-and-error.
We provided a solution, which we believe supports the rationale of traceability and
is applicable for practitioners.
Yet, there remain various open questions, which now list.

\textbf{Gap between trained and inference DNN.}
The process of Section \ref{sec:DL} assumes that trained and inference DNNs have the same input/output types.
This is not always true:
one might train a sub-part of the final network in an unsupervised manner,
to learn features potentially valuable for later supervised training \cite{autoencoder}.
One might also train a DNN on a separate dataset or take an already-trained DNN
and remove the latest layers (the most task-specific ones) to adapt the DNN to the targeted functionality.
In such cases, intermediate steps are not connected to the final task
and thus not traceable in the sense considered so far.

\textbf{Dataset.}
An important aspect is the evolution of the dataset:
we assumed so far that the dataset is fixed.
In reality, many systems gather new data along their lifetime.
In such cases, one should consider a form of incremental traceability,
i.e., how to trace new data as it comes along?
In particular, one might need to argue why adding a new datum indeed provides additional valuable information.

\textbf{Explainable AI.}
As mentioned from the beginning, we took a pragmatic approach in this paper, 
leaving aside explainable AI, which is currently not mature enough
(see, e.g., \cite{chih-hong-metrics} for a short review).
However, as the field develops, it will be valuable 
to discover new opportunities for relevant traces.

\textbf{Classical AI.}
Various approaches attempt to mix deep learning with expert knowledge \cite{transfer-learning,probprog}
are promising for safety-critical systems because they allow
to control the learning process and hence to argue better about the resulting functionality.
In some sense, one can interpret this as a form of explainability-by-design.
Even though they do not reflect the state of the practice,
it would be valuable to consider how to trace artifacts added by these methods.

\textbf{Intellectual property.}
In automotive or avionics, the development of the system is distributed among various stakeholders.
In such cases, it is also essential that all stakeholders keep their intellectual property.
This can be problematic for our approach to trial-and-error activities which forces practitioners 
to provide artifact evolutions which might reveal their production secrets.

\appendix
\section{Reviews}

This paper was submitted and rejected at SAFECOMP 2019.
To help the reader better assess the contribution of this work, we include the anonymous reviews verbatim
(no grades were provided).

\subsection{Review 1}

This paper addresses the very important problem of using Deep Neural Networks (DNNs) in safety-critical applications. Many people are working on this problem, including so-called “self-explaining AI”. These authors describe a more pragmatic approach that can be applied today, before the other techniques being studied are ready for prime time. It involves applying a classical approach of traceability to the artifacts of deep learning. Arguing that traceability, at its best, forces the developer to explain why things were done (“rationale”) and helps to find design shortcomings (e.g. requirements never implemented), the authors propose that transposing the traceability concept to DNNs could offer some of the same advantages.

In order to develop this concept, it was necessary for the authors to impose some rather severe restrictions, which certainly circumscribe the practical usefulness of their ideas. On the other hand, it permits them to discuss some interesting possibilities for immediate application of the traceability technique already today. They present ideas about what the artifacts might be in the case of DNNs. They deal with the problem of the fact that DNNs today basically evolve by trial and error – but face it head-on with proposals for handling this, including for example the idea of a metric that must justify how the evolution of the DNN was “better” than the previous one.

Overall, I believe that the paper represents an interesting attempt to see what traceability of DNNs could look like when applied to DNNs as they stand today. The authors honestly list a number of obstacles and questions that must be answered in the future, but I believe that the paper could generate some useful discussions at SAFECOMP.

\subsection{Review 2}

Initial ideas and work on traceability of DNNs for safety-critical applications. Strong limitation of scope. Further work is needed for application
Chapter 2 provides an compact but excellent summary of recent, related work on the subject. The list of references is very comprehensive.
Chapter 3 is a well summary on the motivation (mostly linked to DO178C) and approaches of requirements traceability. It needs to mentioned that not only DO178C but also ISO26262 do not enforce the order in which software artefacts are created during development. It would have been worth mentioning clause 7 and figure 9 of ISO26262-10 2nd Ed.
chapter 4 describes the step by step approach on training a DNN.
Chapter 5 aims on mapping the approach described in chapter 3 to the DNN development as described in chapter 4. It newly proposes a "domain coverage model" as a refinement of high level requirements to trace from HLR to the training data set. But at the end this does not give any confidence that a trained DNN will fulfill its HLR. The chapter is a very good discussion on the problems that occur when transforming the idea traceability to the development of DNNs. The final proposal to trace all versions of the network (generated during training) together with the increment in the metric value is a excellent combination of design and testing traceability.

Ref. 14 shall be corrected: "ISO 26262: Road vehicles ". Please check for any updates of the paper w.r.t 2nd Ed. from 2018.

\subsection{Review 3}

The paper introduces an approach for tracing requirements to code if DNNs are used instead of conventional code. The authors introduce a concept they call a domain coverage model replacing low-level requirements and they recommend to trace the developers' decisions during the trial-and-error-cycles instead of architectural/structural code properties.

The paper is sometimes quite difficult to read so that is not clear what the authors actually want to express. It should definitely be proof-read by a native speaker. Overall, the paper leaves a rather immature impression - content-wise and presentation-wise.

I'm not completely sure about the actual contribution of the paper. The authors self-confidently explain that all the related work is rather immature and not usable for their purpose. After that, they provide a brief introduction to traceability in general. Than they illustrate the differences between a conventional development process and the process used for creating DNN-based software. Finally, they eventually come to what could be the contribution: They recommend to use a domain coverage model instead of low-level-requirements. This model looks quite similar to models describing operational situations such as ODD or approaches used for deriving test scenarios, but according to the authors it seems to be different. Unfortunately, they do not provide more than about half a page roughly describing the general idea.  Instead of tracing to weights, training sets etc., the authors also recommend to trace to the developers' decisions, i.e. to a rationale why and how the changes!
  of the net lead to better results. Again, there is not more than about half a page roughly explaining what the authors mean. They furthermore explain that this approach requires a metrics for measuring what "better" actually means. But there a lot of "shoulds", i.e., how the metric should be, rather than concrete suggestions for a concrete metric or even evaluations of their metric.

Overall, the paper looks rather immature. The authors describe a rather rough idea and it remains rather unclear how that idea concretely looks like. Moreover, the paper's presentation requires a lot of polishing.

\bibliography{paper}
\bibliographystyle{splncs04}

\end{document}